\documentclass[letterpaper, 10 pt, conference]{ieeeconf}  

\IEEEoverridecommandlockouts                              

\usepackage{url}
\usepackage{amssymb}
\usepackage{amsmath}
\usepackage{mathrsfs}
\usepackage{algorithm}
\usepackage{graphicx}
\usepackage{pdfpages}
\usepackage{wrapfig}
\usepackage{booktabs}
\usepackage{algpseudocode}
\usepackage{multirow} 
\usepackage{pifont}

\newcommand{\Yes}{\textcolor{green}{\textbf{\ding{51}}}}
\newcommand{\No}{\textcolor{red}{\textbf{\ding{55}}}}

\title{\LARGE \bf TwinAligner: Visual-Dynamic Alignment Empowers Physics-aware Real2Sim2Real for Robotic Manipulation}

\author{Hongwei Fan$^{1,2*}$, Hang Dai$^{1,2*}$, Jiyao Zhang$^{1,2*}$, Jinzhou Li$^{1,2}$, Qiyang Yan$^{1,2}$, \\Yujie Zhao$^{1,2}$, Mingju Gao$^{3}$, Jinghang Wu$^{1,2}$, Hao Tang$^{3}$ and Hao Dong$^{1,2\dag}$%
\thanks{*Equal contribution. \dag Corresponding author. $^{1}$CFCS, School of Computer Science, Peking University; $^{2}$PKU-AgiBot Lab; $^{3}$State Key Laboratory of Multimedia Information Processing,
School of Computer Science, Peking University.}}

\begin{document}
\maketitle
\thispagestyle{empty}
\pagestyle{empty}

\begin{abstract}
The robotics field is evolving towards data-driven, end-to-end learning, inspired by multimodal large models. However, reliance on expensive real-world data limits progress. Simulators offer cost-effective alternatives, but the gap between simulation and reality challenges effective policy transfer. This paper introduces TwinAligner, a novel Real2Sim2Real system that addresses both visual and dynamic gaps. The visual alignment module achieves pixel-level alignment through SDF reconstruction and editable 3DGS rendering, while the dynamic alignment module ensures dynamic consistency by identifying rigid physics from robot-object interaction. TwinAligner improves robot learning by providing scalable data collection and establishing a trustworthy iterative cycle, accelerating algorithm development. Quantitative evaluations highlight TwinAligner's strong capabilities in visual and dynamic real-to-sim alignment. This system enables policies trained in simulation to achieve strong zero-shot generalization to the real world. The high consistency between real-world and simulated policy performance underscores TwinAligner's potential to advance scalable robot learning. Code and data will be released on \url{https://twin-aligner.github.io}.

\end{abstract}

\section{Introduction}

Inspired by advances in multimodal large models~\cite{hurst2024gpt,Qwen2.5-VL}, robotics is transitioning from modular skill learning to data-driven end-to-end learning. Unlike multimodal models that can directly leverage internet data, robotics relies on data from embodied agents interacting with the real world. Recent efforts~\cite{o2024open,contributors2025agibotworld} have focused on creating datasets through the teleoperation of robots in the real world. However, the high costs and uncontrollable data distribution in real-world collections significantly limit the development of data-driven imitation learning.

In parallel, physics-based simulators~\cite{mittal2023orbit,todorov2012mujoco,Genesis} offer a more cost-effective and controllable environment for data collection~\cite{contributors2025agibotdigitalworld}, making them attractive for developing \textbf{Real2Sim2Real} systems that enable rapid experimentation and policy learning. However, transferring policies from simulation to the real world remains challenging due to the \textit{multi-faceted simulation-to-reality (Sim2Real) gap}, spanning differences in visual appearance and dynamics. %
Domain randomization~\cite{yang2025noveldemonstrationgenerationgaussian,yuan2025roboengine} has been explored to bridge this gap, yet it requires extensive data collection and high parameter variability, compromising sample efficiency for improved Sim2Real generalization. Moreover, it fails to provide a simulation environment strictly aligned with the real world for accurate model evaluation~\cite{li2024simpleenv}.

Consequently, \textit{developing a comprehensively aligned Real2Sim2Real system between simulation and the real world} is both highly valuable and technically challenging. Such a system would not only substantially reduce human effort in the data collection process but also enable closed-loop evaluation of robotic policies, thereby shortening the iteration cycle for policy development. Achieving this goal requires addressing two intertwined dimensions of the Sim2Real gap. The visual gap stems from mismatches in rendering fidelity, texture realism, and geometric consistency between simulated and the real world. The dynamic gap arises from approximate physical models and inaccurate physical parameters in simulators.
As evidenced by Tab.~\ref{tab:method_comparison}, despite notable advances in mitigating each gap independently, existing approaches have yet to effectively address both dimensions in a unified manner.

\begin{figure}
    \centering
    \includegraphics[width=0.5\textwidth]{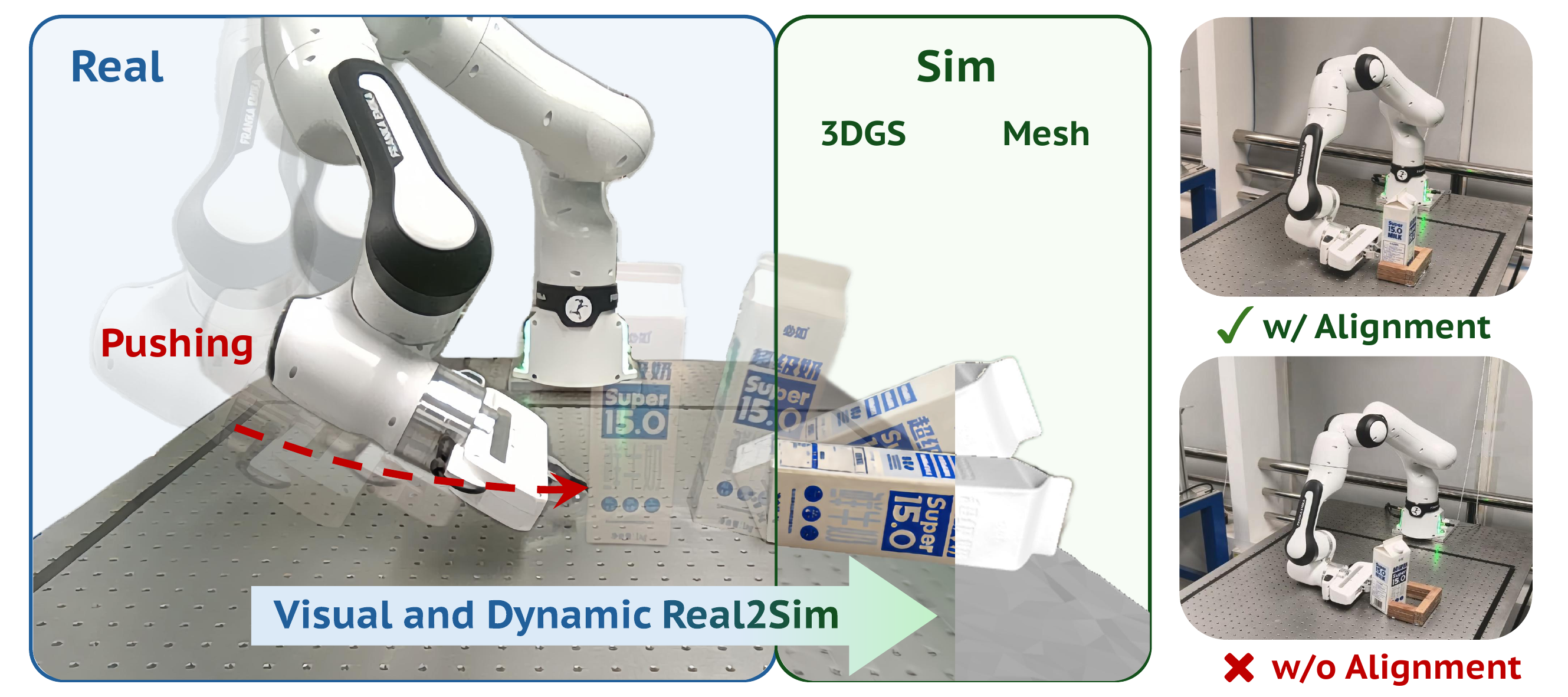}
    \caption{TwinAligner empowers a physics-aware Real2Sim2Real system for policy learning and closed-loop evaluation.}
    \label{fig:teaser}
    \vspace{-2em}
\end{figure}

\begin{table*}[t]
    \centering
    \caption{Comparison of design with other Real2Sim2Real methods. The abbreviations of ``Digital Twin'': O-object, S-scene, R-robot.}
    \resizebox{1.0\textwidth}{!}{
    \begin{tabular}{lccccccc}
    \toprule
    \textbf{Methods}  & \textbf{Real2Sim2Real} & \textbf{Digital Twin} & \textbf{Viewpoint Alignment} & \textbf{Dynamic Alignment} & \textbf{Gradient-free Dynamics} \\
    \midrule
    RialTo~\cite{villasevil2024reconciling} & \Yes & OS & \No & \No & \No  \\
    SplatSim~\cite{qureshi2024splatsimzeroshotsim2realtransfer} & \Yes & OSR  & \No & \No & \No \\
    Re3Sim~\cite{han2025re3sim}                              & \Yes & OS            & \Yes & \No      & \No \\
    Scalable Real2Sim~\cite{pfaff2025_scalable_real2sim}     & \No   & OR            & \No   & \No      & \Yes \\
    PIN-WM~\cite{pin-wm}                                     & \Yes & O                   & \No   & \Yes    & \No \\
    \midrule
    \textbf{TwinAligner (Ours)}                                            & \Yes & OSR     & \Yes & \Yes    & \Yes \\
    \bottomrule
    \end{tabular}
    }
    \label{tab:method_comparison}
\end{table*}

In this paper, we introduce \textbf{TwinAligner}, a novel unified Real2Sim2Real system designed to address both visual and dynamic alignment challenges. The framework consists of two core components:
\textbf{(1) Visual Alignment Module.} Inspired by the real-time, high-quality, and editable rendering capabilities of 3D Gaussian Splatting (3DGS)~\cite{3dgs}, this module achieves pixel-level alignment between simulation and the real world using editable 3DGS rendering. Furthermore, we reconstruct an SDF-based mesh~\cite{neus} that is aligned with the 3DGS representation, enabling accurate physical interactions. This design provides a reliable, plug-and-play rendering engine compatible with various existing physics engines. As shown in Tab.~\ref{tab:method_comparison}, our method supports Real2Sim alignment for objects, scenes, and robots concurrently.
\textbf{(2) Dynamic Alignment Module.} Aligning dynamics is critical for successful Sim2Real transfer of policies trained in simulation, especially in high-dynamics and non-prehensile scenarios. However, this aspect is often overlooked in existing Real2Sim2Real systems. In TwinAligner, we address the dynamic Sim2Real gap from both the robot dynamics and object dynamics perspectives. Specifically, we align the robot dynamics by closing the gap of joint states, and align the object dynamics by enforcing consistency in real-world and simulated point clouds. 
This dual alignment enables robust physics-aware Real2Sim2Real transfer, even in challenging scenarios involving non-prehensile manipulation.

Overall, TwinAligner brings two key advancements to the field of robot learning by rapidly constructing a digital twin highly aligned with the real world visually and dynamically:
\begin{itemize}
    \item \textbf{Efficient Data Collection:} It provides an economical and efficient method for embodied data collection, reducing the gap between simulation and real-world data, and enhancing the sample efficiency of simulation data.
    \item \textbf{Reliable Iterative Cycle:} It establishes a reliable closed-loop system for data collection, model training, and model evaluation, accelerating the iteration speed of robot learning algorithms.
\end{itemize}

To validate our method, we conduct quantitative evaluations of TwinAligner's Real2Sim capabilities, focusing on both visual and dynamic aspects. Fig.~\ref{fig:teaser} illustrates our method's alignment capabilities using a dynamic pushing example. Furthermore, to demonstrate the benefits of TwinAligner's high-quality alignment in robot learning, we perform Real2Sim2Real experiments across various robot manipulation tasks. The results indicate that TwinAligner can achieve zero-shot generalization to real-world scenarios without additional real data, maintaining sample efficiency comparable to real-world trajectories. Additionally, the experiments reveal a high consistency in policy performance between the real world and simulation created by TwinAligner, underscoring its potential to significantly enhance scalable robot learning research.

\section{Related Works}
\label{sec:related_works}

\begin{figure*}[t]
    \centering
    \includegraphics[width=\textwidth]{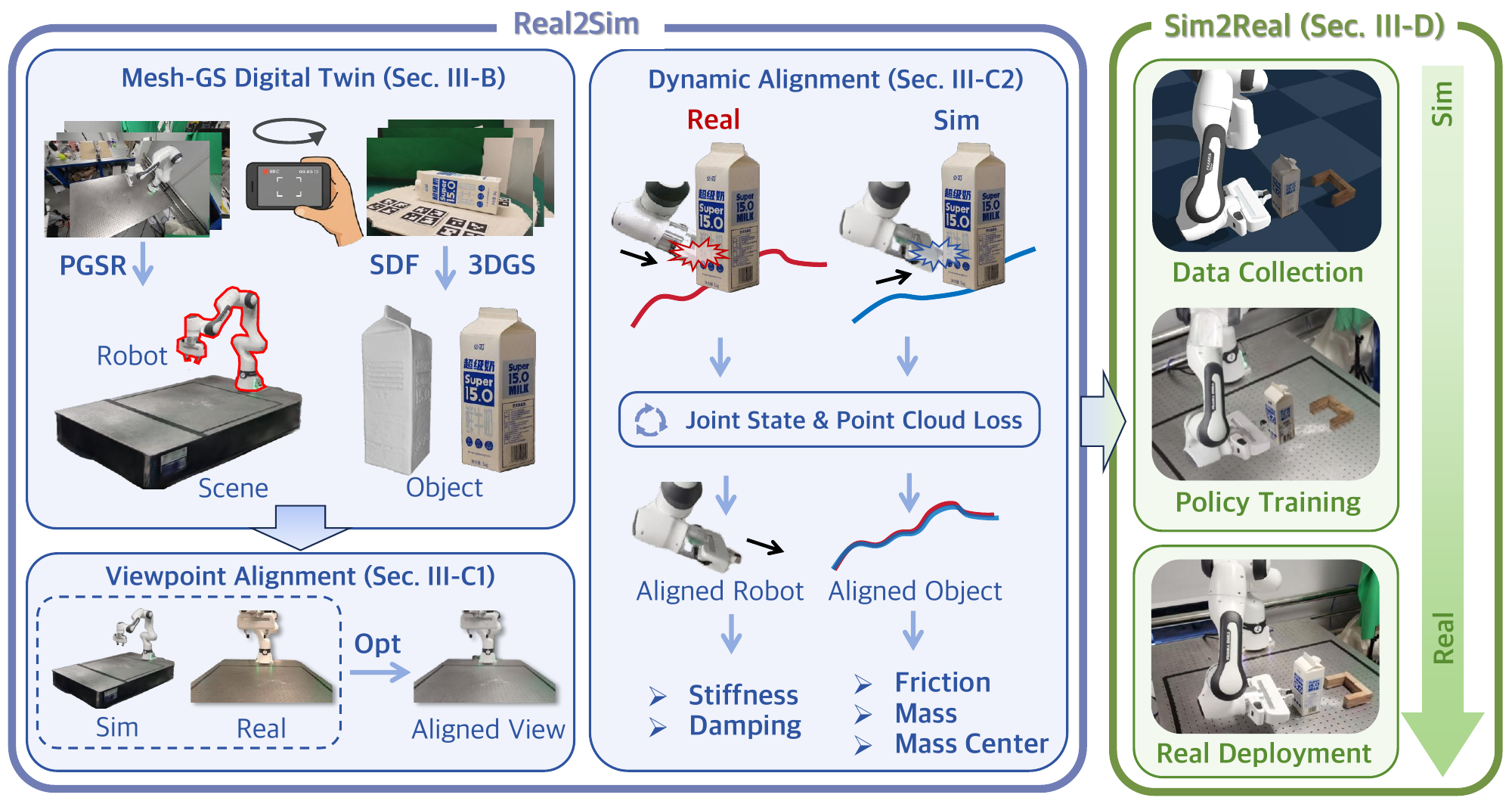}
    \caption{\textbf{Overview of TwinAligner}. The framework consists of two phases: Real2Sim and Sim2Real. In the Real2Sim phase, both Visual Alignment and Dynamic Alignment between the simulation and the real world are considered. Policies trained on robotic trajectories collected in the simulation created through Real2Sim can directly zero-shot generalize to the real world.}
    \label{fig:framework}
\end{figure*}

\subsection{Real2Sim2Real}
    For efficient and low-cost data collection and policy evaluation in simulation, the Real2Sim2Real paradigm aims to build a closed-loop alignment system to mitigate the difference between real and simulated environments. RialTo~\cite{villasevil2024reconciling} establishes a Real2Sim2Real system to fine-tune real-world imitation learning through reinforcement learning in Real2Sim environments. Following RialTo, many works~\cite{qureshi2024splatsimzeroshotsim2realtransfer,han2025re3sim,jia2024discoverse,li2024robogsimreal2sim2realroboticgaussian} focus on improving the visual quality of Real2Sim by employing 3D Gaussian Splatting (3DGS) methods~\cite{3dgs}. However, these methods neglect the dynamic Real2Sim2Real for physics-aware robotic manipulation. PhysTwin~\cite{jiang2025phystwin} and Real-is-Sim~\cite{jad2025realissim} use the MPM-based method to achieve precise dynamic alignment in the visual domain, while restricting the generalization ability. The most related work to ours is PIN-WM~\cite{pin-wm}, which optimizes physical parameters by combining differentiable rigid physics and rendering. However, the differentiable nature makes PIN-WM sensitive to the reconstruction, viewpoint, and relighting errors. Our TwinAligner framework jointly aligns the visual and dynamic gap in one framework, empowering Real2Sim2Real for physics-aware manipulation.

\subsection{Physics-aware Real2Sim}
    For the purpose of closing the dynamics gap in Real2Sim2Real, another line of works focuses on physics-aware Real2Sim. Several works~\cite{shuai2024pugs, zhai2024nerf2physics, lin2025omniphysgs,zhong2024springgaus,zhang2024mingtong-gnn,jad2024embodiedgaussian,xie2024physgaussian, zhang2024physdreamer,feng2024splashing,huang2024dreamphysics} reconstruct 3DGS with particle physics and multi-modal foundation models, which usually mismatch the real-world. The most related work to ours is Scalable Real2Sim~\cite{pfaff2025_scalable_real2sim}, which simultaneously identifies the attributes of real-world physics and reconstructs the object meshes. However, they focus on real-world physical properties in ignorance of the difference between simulated physics and reality. In contrast, our method, aiming to align simulated and real physics, achieving robust Sim2Real performance.

\section{The Proposed Method}

\subsection{Overview}

TwinAligner~(Fig.~\ref{fig:framework}) closes the physics-aware Real2Sim2Real loop through the three steps. First, to accurately replicate the real-world scenario in physics simulator, \textbf{Mesh-GS Digital Twin} (Sect.~\ref{sec:hdt}) jointly reconstructs the detailed visual appearance and geometry, and ensures that the fundamental rendering and collision effects are aligned to the real-world. Then, to make the simulator learn real-world dynamics, \textbf{Visual-Dynamic Real2Sim Alignment} (Sec.~\ref{sec:visual_dynamic}) aligns the real-world dynamics interaction by jointly estimating the camera viewpoint, robot controller, and object rigid physics. Finally, \textbf{Sim2Real Policy Learning} (Sec.~\ref{sec:app}) is performed on the aligned simulation environment with imitation learning policies, which achieves a reduced Sim2Real gap in both regular and physics-aware robot manipulation scenarios.

\subsection{Mesh-GS Digital Twin for Physics Simulation}

\label{sec:hdt}

With the aim of precisely replicating the physics-aware manipulation scenario in the simulator, one must simultaneously achieve high-quality visual rendering and geometry collision. We propose \textbf{Mesh-GS Digital Twin} that supports scene, robot, and object reconstruction with accurate 3DGS and continuous collision mesh in one module.

\textbf{Settings.} We use a smart phone to capture multiview RGB images $I_{obj}$ and $I_{scene}$ for both object and scene assets. Our goal is to recover the detailed 3DGS $G_{obj}$, $G_{scene}$ and the continuous collision meshes $\phi_{obj}$ and $\phi_{scene}$. For the robot, we extract 3DGS $G_{robot}$ from $G_{scene}$, and also do the articulation for $G_{robot}'=\mathbf{W}(G_{robot},J)$ where $\mathbf{W}$ is the forward kinematics function and $J$ are the joint angles.

\textbf{Rigid object.} For $G_{obj}$ and $\phi_{obj}$, on the one hand, to ensure the \textit{simulator-friendly collision}, we reconstruct the watertight meshes $\phi_{obj}$ with SDF representation~\cite{neus,sdfstudio}. Specifically, for spatial point $x_i$ with ray direction $v_i$, we query the SDF value $s_i$ and color $c_i$ from the feature network $F$, and volume-render the image $\widehat{C}^{\text{SDF}}=\sum_{i=1}^{n}T_i\alpha_ic_i$ with color loss $L_c=L_1(\widehat{C}^{\text{SDF}}, I_{obj})$ as the supervision, where $n$ is the number of sample points on the ray. 

On the other hand, to \textit{balance the quality of mesh collision and rendering}, we initialize $G_{obj}$ with the vertices of $\phi_{obj}$, and alpha-blend the 3DGS rendering results $\widehat{C}^{\text{GS}}$ with $L_1$ RGB and SSIM loss as supervision. For regularization, we implement mask loss~\cite{sam2,groundedsam} and monocular depth loss~\cite{sparsenerf, depthanythingv2} to suppress free-space floatings.

\begin{figure}[t]
    \centering
    \includegraphics[width=0.5\textwidth]{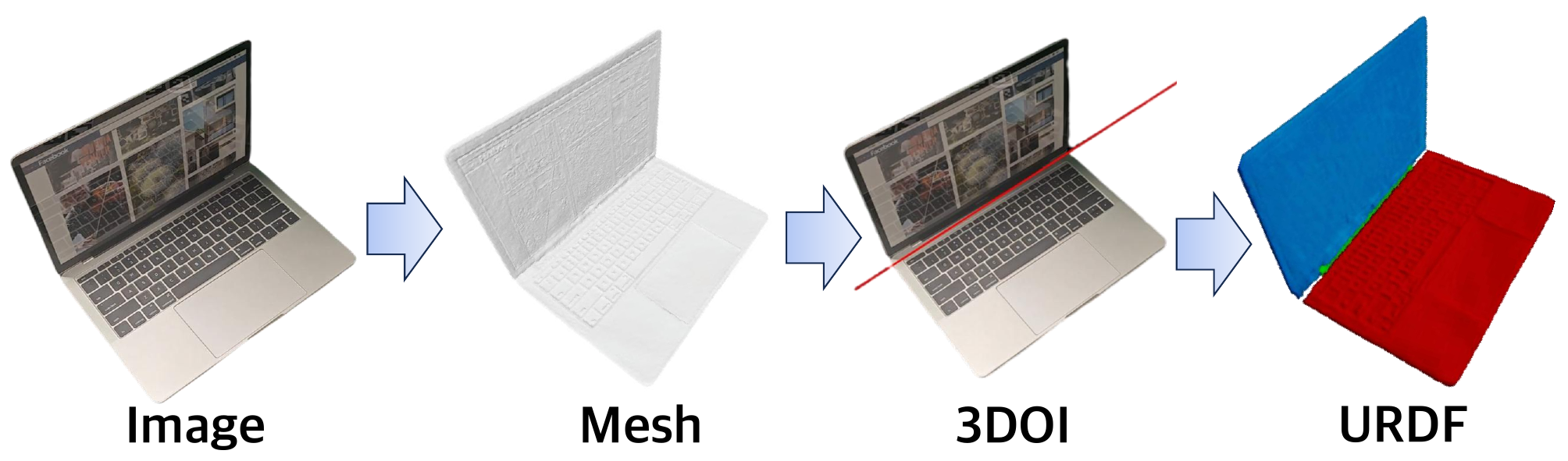}
    \caption{TwinAligner can be easily adapted to articulated object by combining with monocular articulation estimation method 3DOI~\cite{qian2023understanding}.}
    \label{fig:articulation}
\end{figure}

\textbf{Articulated object.} Our digital twin object can be easily adapted to articulated object. Following~\cite{xia2025drawer}, we build an automated articulation labeling pipeline with 3DOI~\cite{qian2023understanding} (Fig.~\ref{fig:articulation}), which uses its annotation results to separate $G_{obj}$ and $\phi_{obj}$ into two parts and estimate the 3D articulation axis with the estimation and depth projection of the 2D joint axis. 

\textbf{Scene and robot.} Regarding the scene, we optimize $G_{scene}$ with planar-based PGSR~\cite{chen2024pgsr} that maintains geometry continuity while maintaining high rendering quality, in which the mesh $\phi_{scene}$ is extracted through TSDF fusion. 

For robot, following SplatSim~\cite{qureshi2024splatsimzeroshotsim2realtransfer}, we align robot URDF $U_{robot}(\cdot)$ and reconstructed $G_{scene}$ with joint angles $J$ by the ICP and k-NN algorithm to obtain the transform $T_{scene2robot}=\text{ICP}(G_{scene}, U_{robot}(J))$ from scene reconstruction to robot URDF, extract robot GS $G_{robot}$ and add forward kinematics. Finally, we convert the real-world object, scene, and robot to the photorealistic and physics simulation-ready assets in the robot axes, and build the basis for the consistent visual-dynamic Real2Sim alignment.

\begin{figure}[t]
    \centering
    \includegraphics[width=0.5\textwidth]{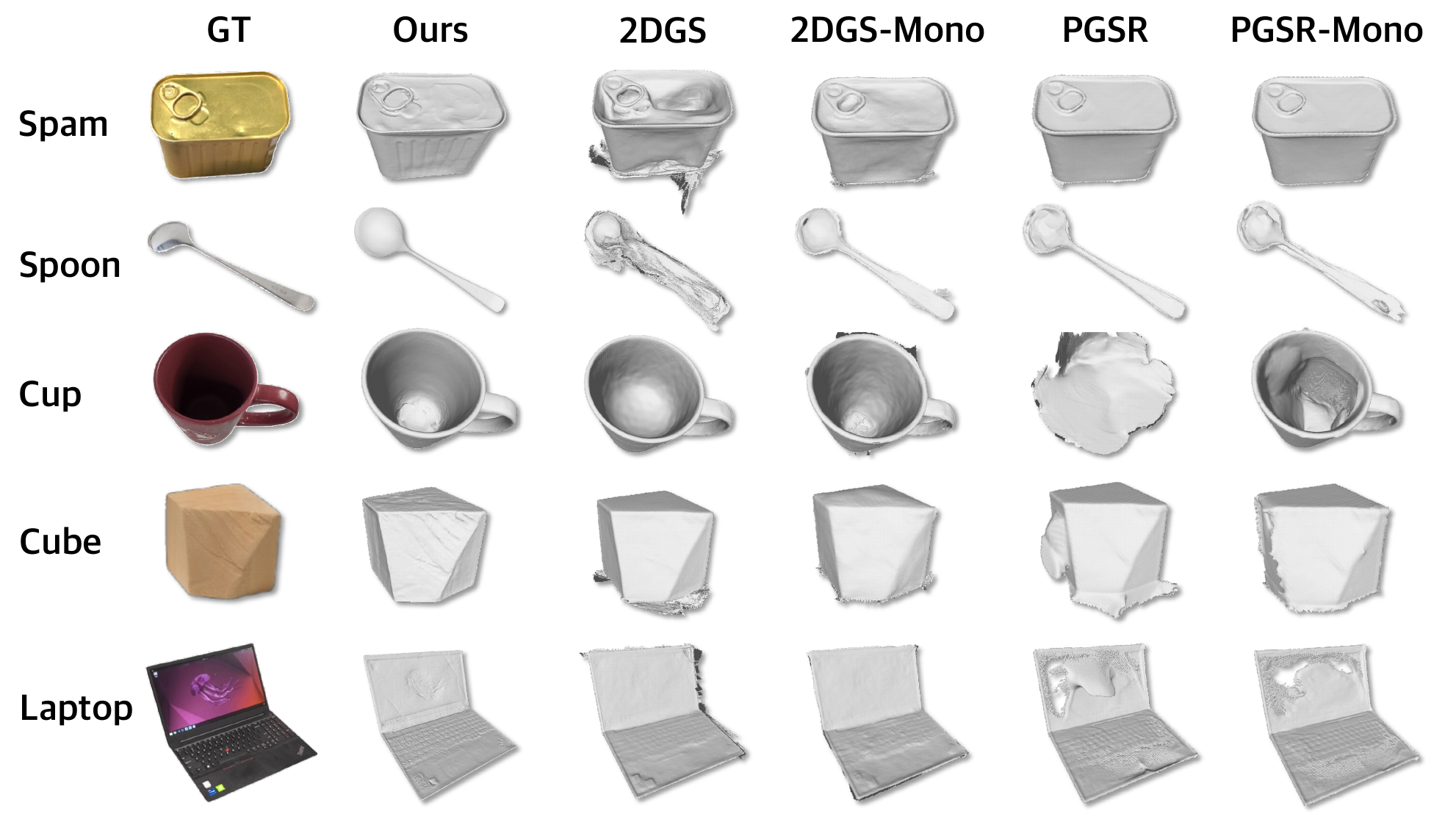}
    \caption{Comparison of geometry reconstruction quality. Our method reconstruct watertight and detailed meshes, while the baseline results contain inaccurate depths, glitches, and holes.}
    \label{fig:mesh_comparison}
    \vspace{-1em}
\end{figure}

\subsection{Visual-Dynamic Real2Sim Alignment}
\label{sec:visual_dynamic}

\begin{figure*}[t]
    \centering
    \includegraphics[width=\textwidth]{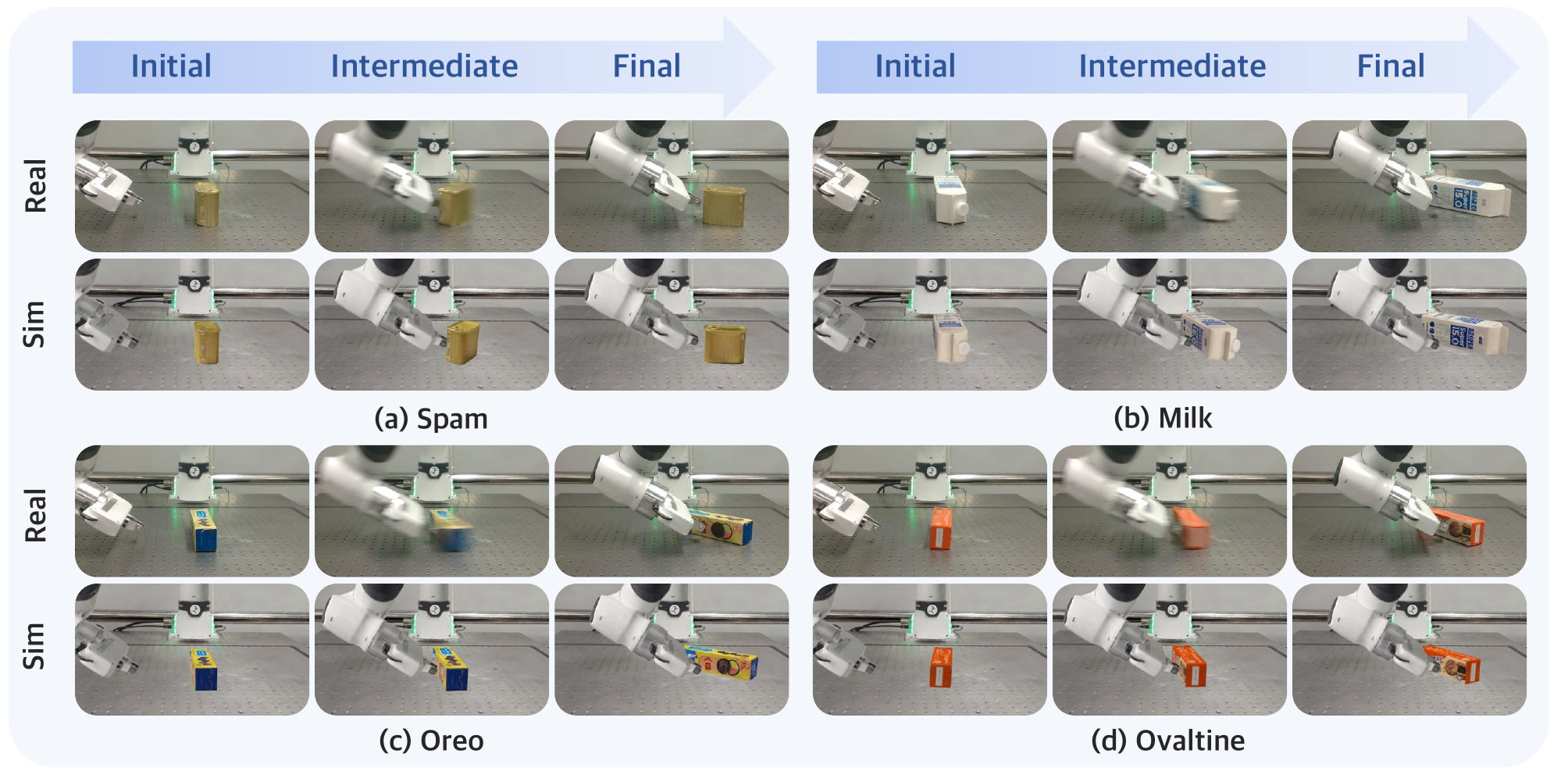}
    \caption{The effectiveness of our visual-dynamic Real2Sim alignment. For the robot-object interaction trajectories, we compare real-world camera observation with physics simulation and 3DGS rendering results from TwinAligner. Our method strictly aligns visual-dynamic gap at the pixel level.}
    \label{fig:traj_render}
    \vspace{-1em}
\end{figure*}

Given simulation-ready digital twin for visual and geometry, we divide the robot simulator to \textit{Dynamic Simulator} and \textit{Camera Renderer}. Set the simulation states $s_i$ and actions $a_i$, viewpoint $p$, rendered image $\widetilde{I}_i$ and dynamics $\theta=\{\theta_{friction}, \theta_{mass}, \theta_{com}, \theta_{robot}\}$ corresponding to friction, mass, center of mass and robot controller. The two procedures for each step can be formulated as
\begin{equation}
\label{eq:sim}
    s_{i+1}=\text{Sim}(s_i, a_i, \phi, \theta)
\end{equation}
\begin{equation}
\label{eq:render}
    \widetilde{I}_i = \text{Render}(s_i, p, G)
\end{equation}
The 3DGS $G$ and the meshes $\phi$ are already built in Sect.~\ref{sec:hdt}. Therefore, the Sim2Real gap remains in \textbf{viewpoint} $p$ and \textbf{dynamics} $\theta$. We detail our alignment methods here.

\textbf{Viewpoint alignment.} The equivalent form of finding the transform from robot to camera $T_{robot2cam}$ is to estimate the pose of robot mesh $U(J)$ in camera observation. We capture an RGB-D image $I'$ with robot joint $J$, and implement the object pose estimator FoundationPose~\cite{wen2024foundationpose}, which predicts the initialized robot-to-camera transformation $T^{coarse}$.

We then utilize an optimizer initialized with $T^{coarse}$ to find the optimal solution $T^{fine}$. Specifically, we render $\widetilde{I}'$ with viewpoint $T^{coarse}$ and 3DGS $G_{scene}$ and $G_{robot}$, and auto-label the silhouettes $M'$ and $\widetilde{M}'$ of the obvious objects or areas (e.g. robot and workspace on the table top) with Segment Anything 2~\cite{sam2} for both $I'$ and $\widetilde{I}'$. We optimize the binary cross-entropy mask matching loss $\mathcal{L}_{\text{BCE}}(M', \widetilde{M}')$ to find the best $T^{fine}$. That is, we assume that the best viewpoint should align the simulated and real image at the pixel level for the obvious image areas. We use the gradient-free optimizer~\cite{kennedy1995particle}, which makes the viewpoint alignment module also adaptable to 3DGS-free simulators.

\begin{table}[tbp]
    \centering
    \caption{Comparison of object rendering quality, measured with PSNR. Higher is better. \textbf{Bold} means the highest metric.}
    \resizebox{0.5\textwidth}{!}{
    \begin{tabular}{lccccc}
    \toprule
    \textbf{Object} & \textbf{Ours} & \textbf{2DGS} & \textbf{2DGS-Mono} & \textbf{PGSR} & \textbf{PGSR-Mono} \\
    \midrule
    Cup    & 35.49 & 36.43 & 32.44 & 22.51 & 29.14 \\
    Spoon  & 35.31 & 31.29 & 30.65 & 31.79 & 22.45 \\
    Laptop & 38.83 & 32.77 & 35.48 & 35.21 & 36.55 \\
    Spam   & 38.17 & 30.53 & 29.23 & 27.47 & 38.25 \\
    Cube   & 42.34 & 45.33 & 29.66 & 44.07 & 43.28 \\
    \midrule
    \textbf{Average} & \textbf{38.03} & 35.27 & 31.49 & 32.21 & 33.93 \\
    \bottomrule
    \end{tabular}
    }
    \label{tab:psnr}
    \vspace{-2em}
\end{table}

\textbf{Dynamic alignment.} After aligning the viewpoint $p$, the only misaligned Real2Sim gap is in rigid physics including friction, mass, center of mass and robot controller parameters $\theta=\{\theta_{friction}, \theta_{mass}, \theta_{com}, \theta_{robot}\}$.

For simultaneously exposing multiple physics attributes such as mass, friction, and center of mass, we propose collecting the dynamic interaction data of the robot and object in a \textbf{ ``Control-Hit-Slide''} manner (Fig.~\ref{fig:traj_render}). \textbf{First}, the robot received control signals and moved towards the target object. \textbf{Then}, the end effector hits the specific part of the object, an instant impulse is applied. \textbf{Finally}, the object slides due to the impulse and slows down with friction. 

In the ``Control'' stage, the robot control $\theta_{robot}$ dominates the Sim2Real gap. We identify the parameters of the robot system following~\cite{chen2023visual} to align the gap, where the robot controller physics $\theta_{robot}$ in the simulation are adjusted to fit the real robot physics $\theta^\prime_{robot}$ as follows:
\begin{equation}
    \mathcal{L}_{robot}=\frac{1}{K}\sum_{i=1}^{K} \|\mathbf{J}(\theta_{robot}, u_i) - \mathbf{J}^\prime(\theta^\prime_{robot}, u_i)\|_2 
\end{equation}
where $\mathbf{J}$ and $\mathbf{J}^\prime$ are the real and simulated PD control functions, which results in the joint positions of the robot in step $i$, $u_i$ is the control signal and $K$ is the number of time steps. After alignment, the gap in robot dynamics is minimized, and the simulated and real robot will ``Hit'' the same position of the object with the same velocity and acceleration. Therefore, the impulse also aligns, making the initial states of the ``Slide'' stage consistent.

The object then `Slides'' on the table. We formulate the physics system as the synergy of the rigid physics parameters of the object $\theta_{friction}, \theta_{mass}$ and $\theta_{com}$:

\begin{equation}
\frac{d\boldsymbol{v}}{dt} = -\theta_{friction} \cdot g \cdot \boldsymbol{e}
\end{equation}
\begin{equation}
\boldsymbol{I} \cdot \frac{d\boldsymbol{\omega}}{dt} = \boldsymbol{r}(\theta_{com}) \times \left( -\theta_{friction} \cdot \theta_{mass} \cdot g \cdot \boldsymbol{e} \right)
\end{equation}

The first and second equations represent the translation and rotation procedures of the object. For translation, $\boldsymbol{v}$, $g$ and $\boldsymbol{e}$ are linear velocity, acceleration of gravity, and unit vector of hitting direction. For rotation, $\boldsymbol{I}$, $\boldsymbol{\omega}$ and $\boldsymbol{r}$ are inertia, angular velocity, and the vector from hitting position to center of mass. This is a complex non-linear system, to solve which we design the optimization method as follows.

To strictly replicate the real-world object pose in simulation and eliminate rendering or relighting errors, for the real-world trajectories, we estimate the initial object 6-DoF poses $\{T_i\}_{i=0}^{K}$ with~\cite{wen2024foundationpose}. In the simulation, we replay the same robot control signals as in the real world, and obtain the corresponding poses $\{\widehat{T}_i\}_{i=0}^{K}$. Then, we employ point cloud-based ADD and ADD-S losses as follows:
\begin{equation}
\mathcal{L}_{obj}=\frac{1}{K}\sum_{i=1}^K\left(
    \mathcal{L}_{\text{ADD}}(T_i, \widehat{T}_i, \phi_{obj})+\mathcal{L}_{\text{ADD-S}}(T_i, \widehat{T}_i, \phi_{obj})
\right)
\end{equation}

We adopt gradient-free optimization~\cite{kennedy1995particle} which ensures compatibility with non-differentiable simulators such as \cite{mittal2023orbit}. Note that our method can be easily adapted from few-shot to many-shot real-world interaction data, and combined with a high-parallel simulation engine to accelerate.

\subsection{Sim2Real Policy Learning}
\label{sec:app}

In the Real2Sim stage, we have aligned all the assets of simulated and real environments, including 3DGS $G=\{G_{obj}, G_{scene}, G_{robot}\}$, meshes $\phi=\{\phi_{obj}, \phi_{scene}, U_{robot}\}$, viewpoint $p$ and rigid physics parameters $\theta$. The strict alignment empowers two main applications for physics-aware Sim2Real manipulation: \textbf{Zero-shot Policy Generalization} and \textbf{Cross-environment Policy Evaluation}.

\textbf{Zero-shot policy generalization}. We perform imitation learning policies Diffusion Policy (DP)~\cite{chi2023diffusionpolicy} and RISE~\cite{wang2024rise} on the TwinAligner engine. For collecting expert demonstration trajectories, we establish tele-operation in the simulation environment, and record the joint angles $J_i$, end-effector state $E_i$, and object state $s_i$ simulated through Eq.~\ref{eq:sim}. Then, we render the camera observation $o_i=\{I_i, D_i\}$ including the RGB image $I_i$ and the depth image $D_i$ through Eq.~\ref{eq:render}. Finally, we train the policies based on the end-effector state $E_i$ and the camera observation $o_i$. Given a sequence of observations $\{o_{i-h+1}, \dots, o_i\}$, the policy $\pi$ predicts a sequence of future robot actions $\{a_{i+1}, a_{i+2}, \dots, a_{i+n}\}$. We adopt pretrained ResNet-18 and sparse 3D encoder~\cite{choy20194d} for DP and RISE.

\textbf{Cross-environment policy evaluation.} Thanks to the aligned environment, we can also evaluate the pre-trained policy in TwinAligner. For time step $i$, the policy model is formulated as $a_i=\pi(o_i)$. Combined with Eq.~\ref{eq:sim} and Eq.~\ref{eq:render}, TwinAligner closes the ``Sim-Render-Policy'' loop. Starting with randomized initial states, the policy interacts with TwinAligner, resulting in the simulated success rates.

\section{Experiments}

\begin{figure*}[t]
    \centering
    \includegraphics[width=\textwidth]{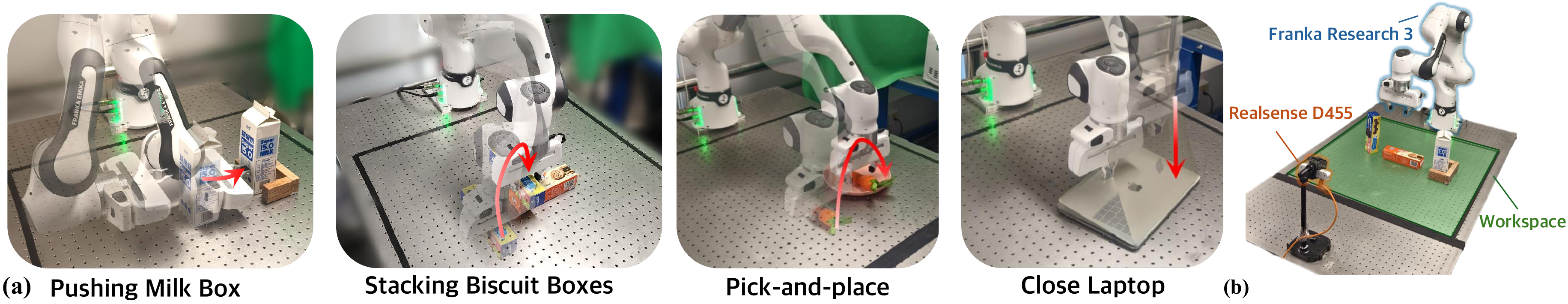}
    \caption{(a) Tasks of policy evaluation, covering from physics-related tasks to regular robotic manipulation tasks. (b) Our hardware setup includes a Franka Research 3 (FR3) robot, a Realsense D455 depth camera and a shock-proof table.}
    \label{fig:tasks}
\end{figure*} 

\begin{table*}[htbp]
\centering
\caption{Success rates of Sim2Real policy learning. \textbf{``-''} means that the method \textbf{does not support} the corresponding task.}
\resizebox{0.9\textwidth}{!}{
\begin{tabular}{lcccccccc}
\toprule
\multirow{2}{*}{\textbf{Settings}} & \multicolumn{2}{c}{\textbf{Pushing Milk Box}} & \multicolumn{2}{c}{\textbf{Stacking Biscuit Boxes}} & \multicolumn{2}{c}{\textbf{Pick-and-place}} & \multicolumn{2}{c}{\textbf{Closing Laptop}} \\
\cmidrule(lr){2-3} \cmidrule(lr){4-5} \cmidrule(lr){6-7} \cmidrule(lr){8-9}
& \textbf{DP} & \textbf{RISE} & \textbf{DP} & \textbf{RISE} & \textbf{DP} & \textbf{RISE} & \textbf{DP} & \textbf{RISE} \\
\midrule
SplatSim + Gemini & 5 / 15 &  3 / 15 & 0 / 15 & 0 / 15 & 5 / 15 & 2 / 15 & - & - \\
PIN-WM & 6 / 15 & 5 / 15 & - & - & - & - & - & -\\
Ours & 10 / 15 & 7 / 15 & 14 / 15 & 7 / 15 & 9 / 15 & 8 / 15 & 10 / 15 & 7 / 15 \\ 
Real2Real & 13 / 15 & 9 / 15 & 13 / 15 & 10 / 15 & 11 / 15 & 10 / 15 & 13 / 15 & 13 / 15 \\
\bottomrule
\end{tabular}}
\label{tab:sim2real}
\vspace{-1em}
\end{table*}

\begin{table}[htbp]
    \centering
    \caption{Comparison of dynamic Real2Sim on our method and PIN-WM. Lower is better. \textbf{Bold} means the lowest metric.}
    \begin{tabular}{lcccc}
    \toprule
    \multirow{2}{*}{\textbf{Object}} & \multicolumn{2}{c}{\textbf{ADD}(cm)} & \multicolumn{2}{c}{\textbf{ADD-S}(cm)} \\
    \cmidrule(lr){2-3} \cmidrule(lr){4-5}
    & \textbf{Ours} & \textbf{PIN-WM} & \textbf{Ours} & \textbf{PIN-WM} \\
    \midrule
    Milk & 1.45 & 2.24 & 0.80 & 1.10  \\
    Oreo  & 1.53 & 4.13 & 0.90 & 2.33  \\ 
    Ovaltine & 1.20 & 1.52 & 0.67 & 0.92  \\ 
    Spam & 1.36 & 1.58 & 0.74 & 0.75   \\
    \midrule
    \textbf{Average} & \textbf{1.39} & 2.37 & \textbf{0.78} & 1.28  \\
    \bottomrule
    \end{tabular}
    \label{tab:physical_alignment}
    \vspace{-2em}
\end{table} 

In this section, we answer the following questions.
\begin{itemize}
    \item Does TwinAligner have high visual-dynamic Real2Sim consistency? (Sec.~\ref{sec:exp_real2sim})
    \item How is the zero-shot Sim2Real generalization ability of policies trained with TwinAligner? (Sec.~\ref{sec:exp_sim2real})
    \item Can TwinAligner be used as a reliable evaluator for model iteration?
    (Sec.~\ref{sec:exp_reliable_eval})
\end{itemize}

\subsection{Real2Sim Consistency}
\label{sec:exp_real2sim}

\textbf{Object digital twin.} To demonstrate the quality of the rendering and geometry of our digital object twin, we evaluate the quality of the rendering with PSNR and visualize the reconstructed meshes. We choose five objects including mug, spoon, laptop, spam and wooden cube, representing concave, reflective, and textureless objects. We compare our method with 2DGS~\cite{Huang2DGS2024} and PGSR~\cite{chen2024pgsr} on the objects. For fair comparison, we also add monocular depth loss following the implementation of \cite{3dgs}, and the corresponding baselines are called 2DGS-Mono and PGSR-Mono.

The results are shown in Tab.~\ref{tab:psnr} and Fig.~\ref{fig:mesh_comparison}. Regarding rendering quality, our method achieves the highest PSNR on average, thanks to the training of 3DGS on mesh reconstruction. For mesh reconstruction, for all the five objects, our method reconstructs detailed and watertight surfaces compared to all the baselines, even if they are equipped with depth regularization. In contrast to existing baselines, the meshes reconstructed by our method exhibit greater completeness. The results demonstrate that our method can achieve both high geometry quality and rendering quality.

\textbf{Dynamic Real2Sim alignment.} We choose four objects with different physical features to evaluate the dynamic Real2Sim alignment. The spam and ovaltine biscuit boxes are with regular rigid physics. However, to increase the difficulty, we \textit{empty the milk box} and \textit{put a wooden cube} on one side of the oreo box to shift its center of mass. 

We collect 20 clips of robot-object interaction data for all four objects, and perform experiments using the rigid physics parameters predicted by our method and PIN-WM~\cite{pin-wm}. We evaluate the alignment of the object trajectories using the ADD and ADD-S metrics~\cite{wen2024foundationpose} to measure how closely the simulation matches the movements of the real world. Tab.~\ref{tab:physical_alignment} and Fig.~\ref{fig:traj_render} present the quantitative and qualitative evaluation results. Both consistently demonstrate that the dynamics learned by TwinAligner achieve better alignment compared to those inferred by PIN-WM. 

There are two possible reasons why our method is superior. First, due to gradient-free optimization and off-the-shelf pose estimation, our method can resist relighting or rendering errors, which is demonstrated by regular objects Ovaltine and Spam. Second, differentiable simulation may be instable when implemented on physically disturbed objects, such as Milk and Oreo.

\subsection{Zero-shot Sim2Real Generalization}
     To validate the effectiveness of the TwinAligner-based Real2Sim2Real system, we performed imitation learning experiments on four tasks (Fig.~\ref{fig:tasks} (a)):.
    
    \begin{itemize}
        \item \textbf{Pushing Milk Box}: The robot is supposed to push an \textit{empty milk box} into a groove. Failure is judged when the box falls down or the robot fails to push the box to the target.
        \item \textbf{Stacking Biscuit Boxes}: The robot needs to put the Oreo biscuit box with a \textit{biased center of mass} on top of the Ovaltine biscuit box. Failure is judged when the Oreo box is unable to stand stably on Ovaltine box.
        \item \textbf{Pick-and-place}: The robot should pick a toy carrot and place it on a pink plate. Failure is judged when the robot fails to pick or place.
        \item \textbf{Closing Laptop}: The robot is supposed to close the lid of the laptop. Failure is judged when the laptop is still open after the policy.
    \end{itemize}

    \textbf{Settings.} Specifically, for these tasks, we collect 50 demonstration trajectories through tele-operation in both the TwinAligner simulation engine and the real world. For baseline, we choose SplatSim~\cite{qureshi2024splatsimzeroshotsim2realtransfer} that focuses on visual Real2Sim2Real and PIN-WM~\cite{pin-wm} with dynamic Real2Sim2Real. For SplatSim, we prompt Gemini~\cite{comanici2025gemini} and initialize the rigid physics parameters to ensure its basic dynamic Sim2Real ability. We reimplement these baselines with tele-operation and policy learning, and also collect 50 demonstrations in their simulators. We use these trajectories to train the Diffusion Policy (DP)~\cite{chi2023diffusionpolicy} and RISE~\cite{wang2024rise}. DP utilizes RGB images and the end effector state as observations, while RISE uses partial point clouds. This approach allows us to thoroughly assess both the visual (including rendering and geometry) and dynamic Sim2Real gaps. Our hardware setup is shown in Fig~\ref{fig:tasks} (b). We mount the Franka Research 3 (FR3) robot and a Realsense D455 camera on a shock-proof table. Genesis~\cite{Genesis} is used as the physics simulation engine.

    \textbf{Results.} Table~\ref{tab:sim2real} presents the success rates of policies trained on 50 trajectories collected from TwinAligner, the real world, and the two baselines. For all tasks, it is evident that, regardless of whether the representation is based on RGB images or point clouds, policies trained in simulation can directly transfer to the real world. Compared with baselines, our method covers all four types of tasks: pushing, stacking, pick-and-place, and articulated object manipulation. In contrast, SplatSim does not include articulated object manipulation, and PIN-WM only supports the pushing task. 
    
    Regarding performance, our method surpasses the two baselines for all tasks, which is consistent with our high visual-dynamic Real2Sim consistency. Although equipped with differentiable rendering and simulation, in the pushing task, PIN-WM only slightly exceeds the baseline SplatSim + Gemini, which is consistent with its unstable dynamic alignment shown in Tab.~\ref{tab:physical_alignment}. Same as pushing milk box, SplatSim also performs worst in pick-and-place and stacking tasks. For pick-and-place, SplatSim uses the standard 3DGS for scene and object digital twins, which fails to balance visual rendering and geometry quality, and the large Sim2Real gap remains. For stacking the biscuit boxes, Gemini fails to accurately identify the correct mass center with only visual information, which causes the SplatSim policy to pick the mass center incorrectly and stack instably.

    \label{sec:exp_sim2real}

\subsection{Cross-environment Policy Evaluation}
    \label{sec:exp_reliable_eval}
    \begin{figure}[t]
        \centering
        \includegraphics[width=0.48\textwidth]{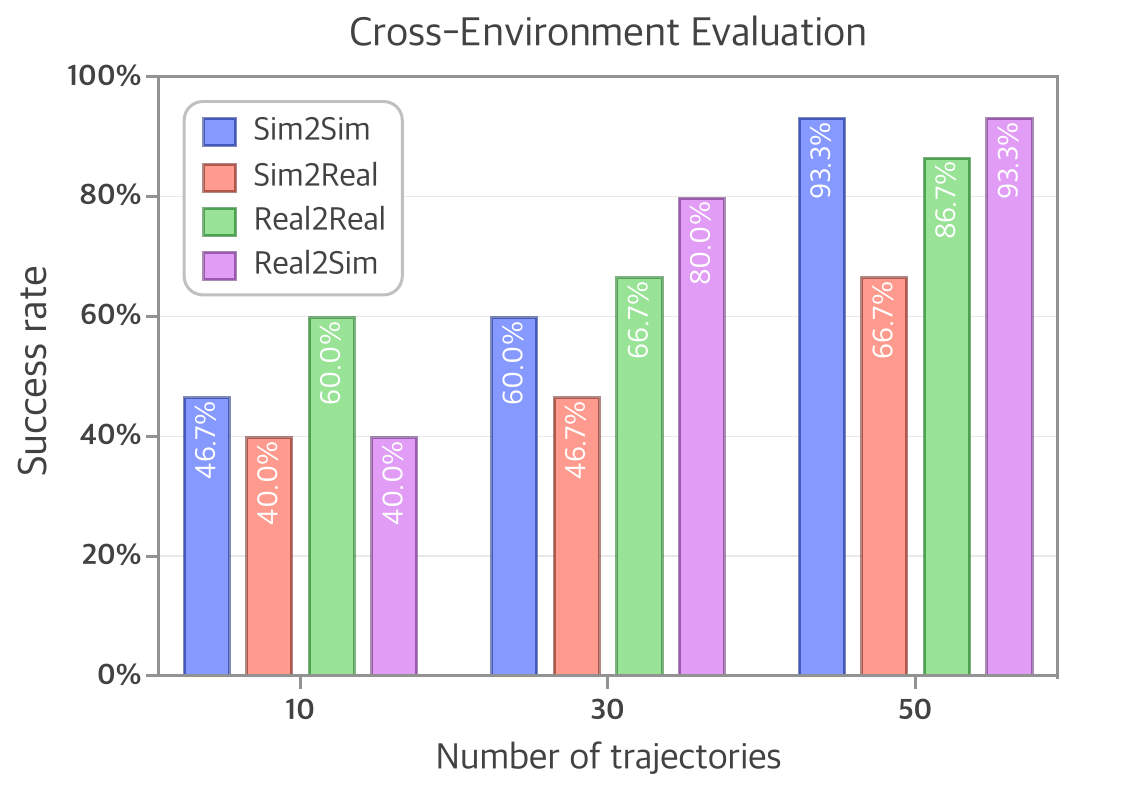}
        \caption{We evaluate TwinAligner across real and simulated environments, which achieves sim-real consistency among different data scales.}
        \label{fig:cross_environments}
        \vspace{-1em}
    \end{figure}
    A well-aligned Real2Sim environment should not only serve the purpose of simulation data collection and enhance the sample efficiency of simulated data but also function as a model evaluator to facilitate rapid iterative optimization of models. We validate TwinAligner's capability in this aspect through cross-environment testing. Specifically, we base on the ``Pushing Milk Box'' task, and test policies trained on both simulated and real data in both simulation and real-world environments. Fig.~\ref{fig:cross_environments} presents the results of the cross-environment validation, showing that although there are minor performance differences, the trend of model performance with respect to the number of training trajectories remains consistent between environments. This demonstrates TwinAligner's potential to establish a reliable closed-loop system for data collection, model training and evaluation, thereby accelerating the iteration speed of robot learning.
    
\section{Conclusion}
\label{sec:conclusion}
In this paper, we propose TwinAligner, a unified Real2Sim2Real framework that closes both the visual and dynamic Real2Sim2Real loops for physics-aware robotic manipulation. To build precise and realistic digital twin assets for physics simulation, we combine SDF reconstruction and 3DGS editable rendering, featuring high accuracy in visual and geometry reconstruction. For jointly aligning the visual-dynamic Sim2Real gap, we propose to estimate viewpoint and optimize simulated dynamics simultaneously in one framework. Experiments demonstrate that TwinAligner achieves not only high alignment quality in visual and dynamics, but also Sim2Real consistency for policy learning.

\section{Limitations}
\label{sec:limitation}

The limitations of our method are threefold. 
First, human efforts for data collection still exist, including the multiview capture of scenes and objects and dynamic trajectory collection. One future direction is to actively reconstruct through physical interaction with the real world. 
Second, the accuracy and speed of dynamic alignment are constrained by the simulation engine. A possible solution to this is to 
learn an NN-based world model that precisely fits robot-object interaction data from the real world.
Third, our dynamic alignment method does not support deformable objects, such as garments. We leave this as future work.

\section{Acknowledgement}
\label{sec:ack}
We would like to thank Xuanyu Lai from Peking University, Han Zhang from Zhejiang University and Haolin Chen from Zhongguancun Academy for their technical supports.

\bibliographystyle{IEEEtran} %
\bibliography{IEEEexample}   %

\end{document}